\documentclass{article}

\usepackage{PRIMEarxiv}

\usepackage[utf8]{inputenc} 
\usepackage[T1]{fontenc}    
\usepackage{hyperref}       
\usepackage{url}            
\usepackage{booktabs}       
\usepackage{amsfonts}       
\usepackage{nicefrac}       
\usepackage{microtype}      
\usepackage{lipsum}
\usepackage{fancyhdr}       
\usepackage{graphicx}       
\usepackage{amsmath, algorithm, algpseudocode, soul} 
\usepackage[super]{nth}
\usepackage{thumbpdf, lmodern}
\usepackage[square,numbers]{natbib}
\usepackage{xcolor}

\graphicspath{{media/}}     

\title{On the development of a practical Bayesian optimisation algorithm for expensive experiments and simulations with changing environmental conditions}

\author{
  Mike Diessner \\
  School of Computing \\
  Newcastle University \\
  Newcastle upon Tyne, UK\\
  \texttt{m.diessner2@newcastle.ac.uk} \\
  \And
  Kevin J. Wilson \\
  School of Mathematics, Statistics and Physics \\
  Newcastle University \\
  Newcastle upon Tyne, UK\\
  \texttt{kevin.wilson@newcastle.ac.uk} \\
  \And
  Richard D. Whalley \\
  School of Engineering \\
  Newcastle University \\
  Newcastle upon Tyne, UK \\
  \texttt{richard.whalley@newcastle.ac.uk} \\
}

\begin{document}
\maketitle

\begin{abstract}
Experiments in engineering are typically conducted in controlled environments where parameters can be set to any desired value. This assumes that the same applies in a real-world setting---an assumption that is often incorrect as many experiments are influenced by uncontrollable environmental conditions such as temperature, humidity and wind speed. When optimising such experiments, the focus should lie on finding optimal values conditionally on these uncontrollable variables. This article extends Bayesian optimisation to the optimisation of systems in changing environments that include controllable and uncontrollable parameters. The extension fits a global surrogate model over all controllable and environmental variables but optimises only the controllable parameters conditional on measurements of the uncontrollable variables. The method is validated on two synthetic test functions and the effects of the noise level, the number of the environmental parameters, the parameter fluctuation, the variability of the uncontrollable parameters, and the effective domain size are investigated. ENVBO, the proposed algorithm resulting from this investigation, is applied to a wind farm simulator with eight controllable and one environmental parameter. ENVBO finds solutions for the full domain of the environmental variable that outperforms results from optimisation algorithms that only focus on a fixed environmental value in all but one case while using a fraction of their evaluation budget. This makes the proposed approach very sample-efficient and cost-effective. An off-the-shelf open-source version of ENVBO is available via the NUBO Python package.
\end{abstract}


\section{Introduction}\label{sec:introduction}
Bayesian optimisation is a sample-efficient optimisation algorithm for the optimisation of expensive-to-evaluate functions that do not possess a mathematical expression or where the expression is too complex to be solved analytically \citep{Mockus1975, Zilinskas1975, Mockus1989, Jones1998, Snoek2012}. Examples of these functions are physical experiments and computer simulations. Simply put, optimisation means finding the optimal parameter values that maximise some objective. In its original form, Bayesian optimisation is a global optimisation algorithm that aims to find a global optimum of a function in a minimum number of function evaluations also called observations. For Bayesian optimisation to be effective, all parameters must be controllable and all environmental factors that could influence the output must remain constant. However, this assumption is only true in completely isolated and controlled environments. Considering more realistic scenarios or experiments where some variables cannot be controlled, it is questionable if this assumption holds. Real environments are generally more complex and environmental conditions, such as humidity, temperature and wind speed, are typically given by uncontrollable external factors.

Many applications of Bayesian optimisation---implicitly or explicitly---assume a simplistic world where all environmental conditions are fixed. In active flow control, for example, where the goal is to control blowing actuators to maximise the reduction of the skin-friction drag over a flat plate, the ambient wind speed is assumed to be fixed \citep{Mahfoze2019, Diessner2022, OConnor2023, Mallor2023}. However, the optimal parameters found from these simulations and experiments give optima for specific wind speeds and cannot necessarily be generalised to other wind speeds. Employing this approach, the experiment requires replication for each wind speed to ensure optimality. Because wind speeds are assumed to be fixed, it is impossible to share observations and thus information between experiments. While observations from different wind speeds will likely not result in the same drag reduction, they will be correlated and contain some information that can be transferred to problems with different wind speeds. Sharing information between different environmental conditions could decrease the number of necessary observations and make Bayesian optimisation more sample-efficient and cost-effective---both essential properties and main objectives of Bayesian optimisation.

Gaussian methods aiming to optimise problems with controllable and environmental variables considered in the past can be mainly classified into two types. The first type aims to find one solution that yields the best results for all realisations of environmental variables. It is assumed that environmental variables take values according to a distribution making the optimum the average expectation over this distribution. Different approaches allow a discrete distribution \citep{Groot2010}, a continuous distribution \citep{Willams2000, Swersky2013} or both \citep{ToscanoPalmerin2018}. \cite{Chang1999, Chang2001} use this type of method to design a femoral component for hip replacements conditional on joint force orientation and cancellous bone prosperity. The aim is to find one design that is optimal for a wide demographic with varying characteristics. The second type of method tries to find multiple solutions for multiple tasks. Thus interest lies not in finding one global optimum but multiple optima, one for each different environmental condition. \cite{Char2019} consider multiple discrete tasks while \cite{Ginsbourger2014} and \cite{Pearce2018} consider continuous environmental variables. 

Although these examples are closely linked to the problem of this article, they have one important distinction. While the environmental values are given externally in real-world applications, it is assumed that they can be set to any desired values in the experiments and simulations above. This deviates from our problem formulation where we explicitly regard problems with uncontrollable environmental variables---in experiments, simulations and the real world. Thus this research is closest related to \cite{Krause2011} who modify upper confidence bound \citep{Srinivas2010} to be suited for optimisation with environmental conditions and derive theoretical bounds for its contextual regret. In contrast to \cite{Krause2011}, this paper considers improvement-based acquisition functions, i.e., expected improvement \citep{Jones1998} and log expected improvement \citep{Ament2023}, give a detailed description of the practical implementation of Bayesian optimisation with environmental conditions, and provide all code at \url{https://github.com/mikediessner/environmental-conditions-BO}. In addition, the reported approach makes less specific assumptions than \cite{Krause2011} who focus on a linear and additive covariance structure for the environmental variables.

This article presents a practical strategy for optimising expensive black-box functions such as physical experiments and computer simulations with influential environmental conditions that are given by external circumstances and cannot be controlled during the optimisation. The strategy extends Bayesian optimisation by (a) fitting a global surrogate model over all controllable and uncontrollable variables, (b) solving the acquisition function conditionally on measurements taken for the uncontrollable variables, and (c) restricting the initial training data that typically consists of many observations generated via a space-filling design to only one observation. It is shown that the proposed algorithm generalises to situations with noisy observations, multiple uncontrollable variables, and uncontrollable variables with different levels of fluctuation and variability. 

To illustrate the value of this approach to the field of engineering, a wind farm simulator is considered with the objective of maximising the annual power generation by finding optimal positions for four wind turbines. The wind direction that affects the power generation significantly is assumed to vary randomly in these simulations representing an influential environmental condition. Results show that ENVBO---the proposed algorithm resulting from the previous investigation---outperforms two other optimisation algorithms used as benchmarks in all but one case. It has the additional benefits of using fewer function evaluations than the benchmarks and can predict wind turbine positions for any possible wind direction within the investigated range. Similar results from the benchmarks could only be achieved by repeating simulation campaigns many times for different wind speeds---an expensive if not infeasible task. Thus the algorithm is sample-efficient and cost-effective and addresses the main problem of expensive black-box function optimisation effectively. An off-the-shelf open-source version of ENVBO is available via the NUBO Python package \citep{Diessner2023} at \url{www.nubopy.com}.

This research article is structured as follows. Section~\ref{sec:methodology} introduces Bayesian optimisation, including surrogate modelling and acquisition functions, and extends it to allow optimisation with changing environmental conditions. Section~\ref{sec:simulations} validates the method on two synthetic test functions---the two-dimensional Levy function and the six-dimensional Hartmann function---and introduces a way of simulating randomly changing environmental conditions via random walks. Section~\ref{sec:empirical} investigates five properties of the proposed method; noise, number of uncontrollable variables, parameter fluctuation, parameter variability and effective domain size (i.e., the actual searched space given by the environmental conditions). Section~\ref{sec:application} considers a nine-dimensional wind farm simulator with eight controllable and one uncontrollable variable. Lastly, a conclusion is drawn in Section~\ref{sec:conclusion}.

\section{Methodology}\label{sec:methodology}
This section provides a brief overview of the fundamentals of Bayesian optimisation, particularly, the surrogate modelling via Gaussian processes and the acquisition functions used to guide the optimisation, before introducing the novel conditional Bayesian optimisation algorithm that enables the optimisation with uncontrollable environmental variables. 

\subsection{Bayesian optimisation} \label{sec:bo}
Consider the $d$-dimensional maximisation problem
\begin{equation} \label{eq:max-problem}
    \boldsymbol x^* = \arg \max_{\boldsymbol x \in \mathcal{X}} f(\boldsymbol x),
\end{equation}
where $\mathcal{X}$ is a continuous input space that is bounded by a hyper-rectangle such that $\mathcal{X} \in [a, b]^d$ with $a, b \in \mathbb{R}$. The objective function $f(\boldsymbol x)$ typically has three properties. First, it is a black-box function that can be provided with an input vector $\boldsymbol x_i$ and allows the observation of the scalar output $y_i$. Beyond this, no other information can be inferred from the function. Second, the function is expensive to evaluate such that significant costs in time, resources or money are generated at each evaluation. Third, the function typically does not possess a derivative or it is too expensive to compute \citep{Frazier2018}. Any noise $\epsilon$ introduced during the evaluation of the objective function $f(\boldsymbol x)$ is assumed to be independent and identically distributed Gaussian noise $\epsilon \sim \mathcal{N} (0, \sigma^2)$ such that an observation can be defined as $y_i = f(\boldsymbol x_i) + \epsilon$. Multiple observation pairs consisting of inputs and outputs are defined as $\mathcal{D}_n = \{(\boldsymbol x_i, y_i)\}_{i=1}^n$. In this article, $\boldsymbol X_n = \{\boldsymbol x_i \}_{i=1}^n$ and $\boldsymbol y_n = \{y_i\}_{i=1}^n$ are used to describe all training inputs and their corresponding outputs.

Bayesian optimisation \citep{Mockus1975, Zilinskas1975, Mockus1989, Jones1998, Srinivas2010, Snoek2012, Shahriari2015,  Frazier2018, Gramacy2020} is an optimisation algorithm based on surrogate modelling with the objective to solve expensive problem~(\ref{eq:max-problem}) in a minimum number of function evaluations. The expensive and opaque nature of these problems requires a sample-efficient optimisation algorithm that keeps costs low to make the optimisation feasible. Bayesian optimisation has emerged as a prime candidate by using a surrogate model $\mathcal{M}$ to represent the unknown objective function $f(\boldsymbol x)$. The mean of the surrogate model's predictive distribution is then used to compute an acquisition criterion $\alpha (\cdot)$ that guides the optimisation process by proposing new input points to be evaluated by the objective function. Bayesian optimisation is a sequential optimisation algorithm that is performed in a feedback loop as illustrated in Algorithm~\ref{alg:bo}. This loop consists of three steps. First, the surrogate model is fitted to the training data $\mathcal{D}_n$. Second, the next candidate point $\boldsymbol x_{n+1}$ is computed by maximising the acquisition criterion $\alpha$. Third, the new candidate point is evaluated by the objective function and its output $y_{n+1}$ is observed. The loop then starts over again adding this newly observed candidate point to the training set. Thus, Bayesian optimisation gathers more information sequentially with each loop. The algorithm stops when a predefined evaluation budget $N$ is exhausted and returns the data pair with the highest observation as its solution.

\begin{algorithm}
\caption{Basic Bayesian optimisation algorithm}\label{alg:bo}
\begin{algorithmic}

\Require Evaluation budget $N$, number of initial points $n_0$, surrogate model $\mathcal{M}$, acquisition function~$\alpha$.

\State Sample $n_0$ initial training data points $\boldsymbol X_0$ via a space-filling design \citep{McKay2000} and gather observations $\boldsymbol y_0$.

\State Set $n = 0$.

\While{$n \leq N -n_0$}

\State Fit surrogate model $\mathcal{M}$ to training data $\mathcal{D}_n = \{ \boldsymbol X_n, \boldsymbol y_n \}$.  
\State Find $\boldsymbol x_{n+1}$ that maximises an acquisition criterion $\alpha$ based on model $\mathcal{M}$, i.e., solve $\max_{\boldsymbol x} \alpha (\boldsymbol x)$.  
\State Evaluate $\boldsymbol x_{n+1}$, observing $y_{n+1}$  
\State Increment $n$.

\EndWhile

\State \Return Point $\boldsymbol x^*$ with highest observation $y^*$.
\end{algorithmic}
\end{algorithm}

\begin{figure}
    \centering
    \includegraphics[width=\linewidth]{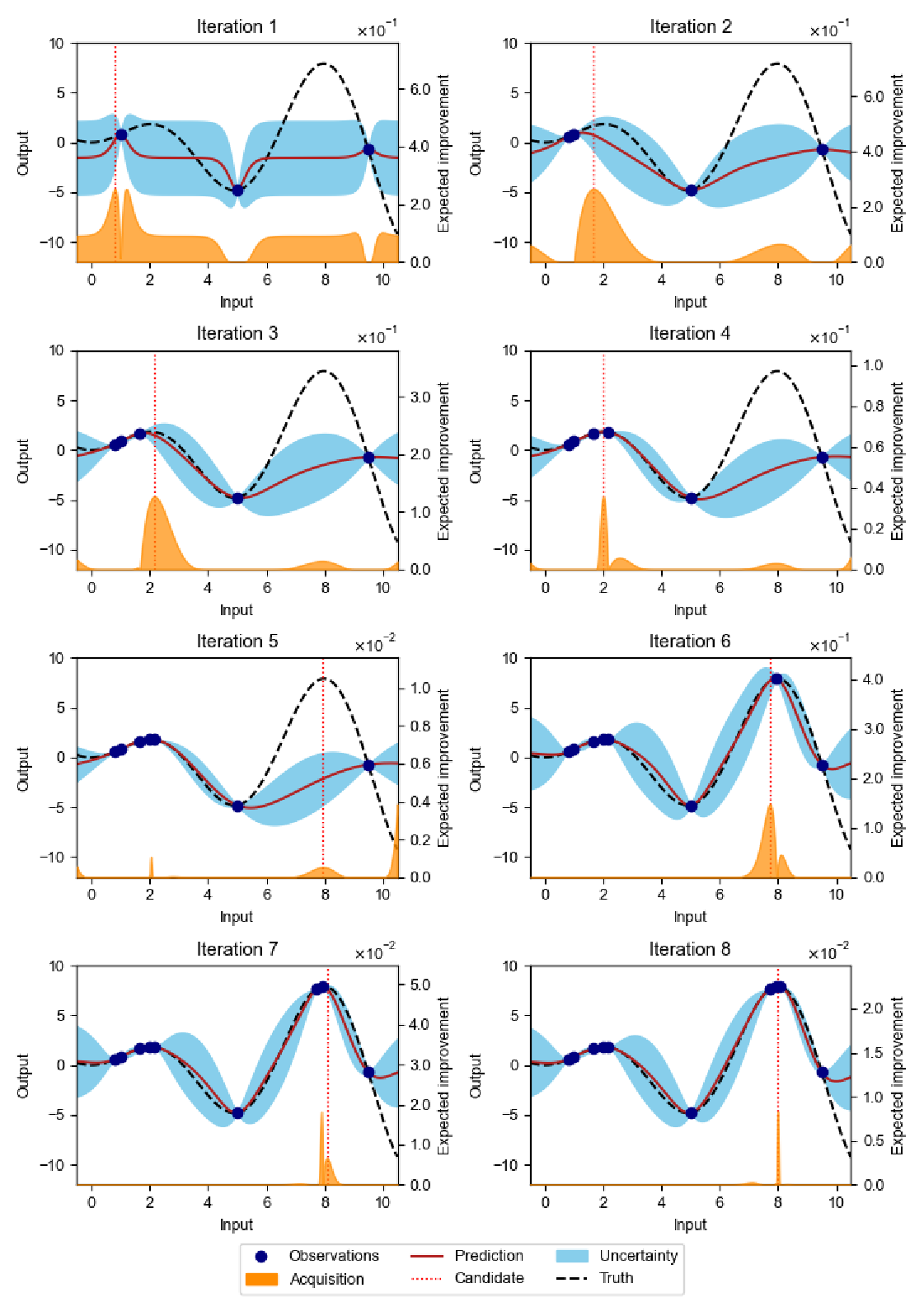}  
    \caption{Bayesian optimisation applied to a 1-dimensional function with one local and one global maximum. Expected improvement is used as the acquisition function. The input space is bounded by $[0, 10]$}
    \label{fig:bo}
\end{figure}

Algorithm~\ref{alg:bo} is illustrated in Figure~\ref{fig:bo}. The objective function (dashed line) is a simple one-dimensional function with one local optimum and one global optimum at $x = 8$. The algorithm is initialised with three observations (dark blue dots) and the surrogate model is fitted providing its prediction (red line) and the corresponding uncertainty (blue area) in the form of 95\% confidence intervals around the prediction. This model is used to compute the acquisition criterion (orange area)--- in this case, expected improvement (see Section~\ref{sec:bo-acq})---that when maximised provides the next candidate point (dashed red line) to be evaluated from the objective function. Bayesian optimisation is run for eight iterations and the surrogate model is updated with each newly evaluated candidate point until the algorithm finds the optimum at iteration six.

\subsubsection{Gaussian process}\label{sec:gp}
While there are alternatives, Gaussian processes \citep{Rasmussen2006, Gramacy2020} are typically selected for the surrogate model $\mathcal{M}$ as they are very flexible and can represent a large number of objective functions $f(\boldsymbol x)$. A Gaussian process is a non-parametric regression model that returns a prediction and its corresponding uncertainty for an unobserved point. Mathematically, it is a finite collection of random variables that have a joint Gaussian distribution. A Gaussian process only requires a prior mean function $\mu_0(\boldsymbol x) : \mathcal{X} \mapsto \mathbb{R}$ and a prior covariance kernel  $\Sigma_0(\boldsymbol x, \boldsymbol x')  : \mathcal{X} \times \mathcal{X} \mapsto \mathbb{R}$ to be fully defined. The mean vector $m(\boldsymbol X_n) := \mu_0(\boldsymbol X_n)$ and the $n \times n$ covariance matrix $K(\boldsymbol X_n, \boldsymbol X_n) := \Sigma_0 (\boldsymbol X_n, \boldsymbol X_n)$ specify the multivariate normal distribution, also called the prior distribution, as
\begin{equation} \label{eq:gp-prior}
    f(\boldsymbol X_n) \sim \mathcal{N} (m(\boldsymbol X_n), K(\boldsymbol X_n, \boldsymbol X_n)).
\end{equation}
This study follows \cite{Snoek2012} and chooses the constant mean function given in Equation~(\ref{eq:const-mean}) as the prior mean function $\mu_0(\cdot)$ and the Mat\'ern $\frac{5}{2}$ kernel presented in Equation~(\ref{eq:matern}) as the prior covariance function $\Sigma_0(\cdot, \cdot)$. The latter is especially suited for the optimisation of real-world problems due to its ability to represent less smooth functions. The Mat\'ern $\frac{5}{2}$ kernel uses the distance between inputs $r = \rvert \boldsymbol x - \boldsymbol x' \lvert$ to compute the uncertainty around its prediction. The kernel uses the output-scale $\sigma^2_f$ to scale the covariance, where smaller values correspond to a smaller deviation from its mean. The characteristic length-scale $l$ quantifies the extent to which values are correlated when moving along the input axes. Smaller length-scales $l$ mean shorter correlation lengths and more variable functions, while larger length-scales $l$ correspond to longer correlation lengths and more constant functions \citep{Gramacy2020, Rasmussen2006}.
\begin{equation} \label{eq:const-mean}
    \mu (\boldsymbol x) = c
\end{equation}
\begin{equation} \label{eq:matern}
    \Sigma_{\text{Mat\'ern}} (\boldsymbol x, \boldsymbol x') = \sigma^2_f \left( 1 + \frac{\sqrt{5}r}{l} + \frac{5r^2}{3l^2} \right) \exp \left( -\frac{\sqrt{5}r}{l} \right)
\end{equation}
The radial basis function kernel given in Equation~(\ref{eq:rbf}) is another popular alternative for the covariance function. It is much smoother than the Mat\'ern kernel and thus is only suited for Bayesian optimisation when it can be assumed that the underlying objective function is necessarily smooth \citep{Snoek2012}.
\begin{equation} \label{eq:rbf}
    \Sigma_{\text{RBF}} (\boldsymbol x, \boldsymbol x') = \sigma^2_f \exp \left( -\frac{r^2}{2l^2} \right)
\end{equation}
Covariance kernels can use one characteristic length-scale $l$ for all input dimensions $d$ or $d$ length-scales $\boldsymbol l = \{l_i\}_{i=1}^d$, one for each input dimension $d$. While the former has less computational overhead due to only requiring one length-scale for all dimensions, it is less flexible as it assigns the same correlation to each dimension. The latter considers each dimension on its own and assigns an individual length-scale to each dimension. This is known as automatic relevance determination (ARD) \citep{Neal1996} as the Gaussian process assigns larger length-scales to more constant dimensions, making them less influential in the computation of the covariance matrix. In contrast, inputs that are variable and change quickly are assigned smaller length-scales, increasing their importance when computing the covariance kernel as changes in these inputs generally affect the prediction significantly. Generally, the inverse of the length-scales $\boldsymbol l$ indicates the relevance of the corresponding input \citep{Rasmussen2006}.

The Gaussian process described above has some hyper-parameters $\theta$ that can be estimated from the training data by maximising the log-marginal likelihood in Equation~(\ref{eq:gp-ll}) via maximum likelihood estimation (MLE) \citep{Rasmussen2006}.
\begin{multline} \label{eq:gp-ll}
    \log p(\boldsymbol y_n \mid \boldsymbol X_n) = -\frac{1}{2} (\boldsymbol y_n - m(\boldsymbol X_n))^\top [K(\boldsymbol X_n, \boldsymbol X_n) + \sigma^2_y I]^{-1} (\boldsymbol y_n - m(\boldsymbol X_n)) \\
    - \frac{1}{2} \log \lvert K(\boldsymbol X_n, \boldsymbol X_n) + \sigma^2_y I \rvert - \frac{n}{2} \log 2 \pi
\end{multline}
 Besides the constant $c$ in the mean function, the signal variance $\sigma^2_f$ and characteristic length-scales $\boldsymbol l$ in the covariance kernel, there is the noise variance $\sigma^2_y$ that reflects the noise level $\epsilon$ that is introduced independent of the objective function $f(\boldsymbol x)$, such that $\theta = \{ c, \sigma^2_f, \boldsymbol l, \sigma^2_y \}$. 
 
The Gaussian process can be used to make predictions along with corresponding uncertainty quantification by computing the posterior predictive distribution~(\ref{eq:gp-posterior}). For $n_*$ test points $\boldsymbol X_*$, it can be computed as the multivariate normal distribution conditional on the training data $\mathcal{D}_n$
\begin{equation} \label{eq:gp-posterior}
    f(\boldsymbol X_*) \mid \mathcal{D}_n, \boldsymbol X_* \sim \mathcal{N} \left(\mu_n (\boldsymbol X_*), \sigma^2_n (\boldsymbol X_*) \right)
\end{equation}
\begin{equation} \label{eq:gp-posterior-mean}
    \mu_n (\boldsymbol X_*) = K(\boldsymbol X_*, \boldsymbol X_n) \left[ K(\boldsymbol X_n, \boldsymbol X_n) + \sigma^2_y I \right]^{-1} (\boldsymbol y - m (\boldsymbol X_n)) + m (\boldsymbol X_*)
\end{equation}
\begin{equation} \label{eq:gp-posterior-cov}
    \sigma^2_n (\boldsymbol X_*) = K (\boldsymbol X_*, \boldsymbol X_*) - K(\boldsymbol X_*, \boldsymbol X_n) \left[ K(\boldsymbol X_n, \boldsymbol X_n) + \sigma^2_y I \right]^{-1} K(\boldsymbol X_n, \boldsymbol X_*),
\end{equation}
where $m(\boldsymbol X_*)$ is the mean vector of length $n_*$ over all test inputs and $K(\boldsymbol X_*, \boldsymbol X_n)$, $K(\boldsymbol X_n, \boldsymbol X_*)$, and $K(\boldsymbol X_*, \boldsymbol X_*)$ are the  covariance matrices of sizes $n_* \times n$, $n \times n_*$, and $n_* \times n_*$ between the training inputs $\boldsymbol X_n$ and the test inputs $\boldsymbol X_*$ respectively.

\subsubsection{Acquisition functions} \label{sec:bo-acq}
Acquisition functions guide the sequential selection of candidate points by quantifying if a certain input point is likely to be a good new candidate point and thus should be evaluated by the objective function $f(\boldsymbol x)$. This is achieved by computing and maximising an acquisition criterion $\alpha (\cdot)$ based on the posterior distribution of the Gaussian process and the available training data $\mathcal{D}_n$. The exact form of the criterion is dependent on the individual acquisition function, however, most acquisition functions have one property in common---the exploration-exploitation trade-off \citep{Shahriari2015, Frazier2018}. Exploration can be defined as choosing candidate points from areas with high uncertainty, that is areas where no training data points were observed and thus little information is available. Exploitation on the other hand is defined as selecting candidate points from areas with a high predictive mean, that is points that are close to high training data points. To understand the importance of balancing exploration and exploitation, consider the extreme cases of pure exploration and pure exploitation. For the former, only points with the highest uncertainty would be selected. While this minimises the uncertainty of the Gaussian process, it is not a sample-efficient approach as information about high predictive means is disregarded and areas with the best-observed outputs are avoided. For the latter, the algorithm will blindly follow the best-performing points and will never explore other areas. The algorithm will probably converge towards the first optimum it discovers making it prone to getting stuck in a local optimum. Thus, a hybrid solution that uses a mixture of exploration and exploitation is beneficial.

This study focuses on two acquisition functions, expected improvement (EI) \citep{Jones1998} and upper confidence bound (UCB) \citep{Srinivas2010}. Expected improvement \citep{Jones1998} is an improvement-based acquisition function that aims to find candidates that perform better than a defined target, typically the best available training data point. Expected improvement is defined as
\begin{equation} \label{eq:acq-ei}
    \alpha_{\text{EI}} (\boldsymbol X_*) = \left(\mu_n(\boldsymbol X_*) - y^{best} \right) \Phi(z) + \sigma_n(\boldsymbol X_*) \phi(z),
\end{equation}
where $z = \frac{\mu_n(\boldsymbol X_*) - y^{best}}{\sigma_n(\boldsymbol X_*)}$, $\mu_n(\cdot)$ and $\sigma_n(\cdot)$ are the mean and the standard deviation of the Gaussian processes' posterior predictive distribution~(\ref{eq:gp-posterior}), $y^{best}$ is the current best observation, and $\Phi (\cdot)$ and $\phi  (\cdot)$ are the cumulative distribution function and probability density function of the standard normal distribution $\mathcal{N}(0, 1)$.

Although expected improvement can never be non-positive mathematically, it can become $0$ numerically when computed due to the floating point precision of the programming language. This results in flat areas where the expected improvement is $0$ and cannot be optimised correctly. To prevent numerically vanishing values, log expected improvement (LogEI) was proposed by \cite{Ament2023} as
\begin{equation} \label{eq:acq-logei}
    \alpha_{\text{LogEI}} (\boldsymbol X_*) = \log_h \left( \frac{\mu_n (\boldsymbol x) - y^{best}}{\sigma_n (\boldsymbol x)} \right) + \log\left( \sigma_n (\boldsymbol x) \right),
\end{equation}
where
\begin{equation}
    \log_h (z) = \begin{cases}
      \log\left( \phi(z) + z \Phi(z) \right) & \text{if} z > -1\\
      -z^2/2 - c_1 + \texttt{log1mexp} \left( \texttt{logerfcx} \left( -z/\sqrt{2} \right) \lvert z \rvert + c_2 \right) & \text{if} z \leq -1
    \end{cases} ,
\end{equation}
where $c_1 = \log(2\pi)/2$, $c_2 = \log(\pi/2)/2$, and $\texttt{log1mexp}$ and $\texttt{logerfcx}$ are stable implementations of $\log\left(1 - \exp(z) \right)$ and $\log\left(\exp(z^2)\text{erfc}(z) \right)$ respectively, where erfc is the complementary error function.

The upper confidence bound (UCB) \citep{Srinivas2010} is an optimistic acquisition function and assumes the uncertainty of the posterior Gaussian process to be true to a predefined level. It can be computed as
\begin{equation} \label{eq:acq-ucb}
    \alpha_{\text{UCB}} (\boldsymbol X_*) = \mu_n(\boldsymbol X_*) + \sqrt{\beta} \sigma_n(\boldsymbol X_*),
\end{equation}
where $\beta$ is a predefined trade-off parameter that can be set for each iteration of the Bayesian optimisation algorithm. This means that it can be kept constant for the full optimisation campaign or be varied at each iteration \citep{Srinivas2010}. \cite{Srinivas2010} investigated some theoretical properties of $\beta$, while \cite{Diessner2022} investigated how different values for $\beta$ affect the optimisation. As the acquisition functions in this study are deterministic, they can be maximised with a deterministic optimiser, such as L-BFGS-B \citep{Zhu1997}.

\subsection{Changing environmental conditions}\label{sec:env-bo}
The Bayesian optimisation algorithm given in Algorithm~\ref{alg:bo} assumes that all parameters that influence the output can be controlled. However, when optimising physical experiments, in many cases there will be variables present that influence the output but cannot be controlled. This article refers to these uncontrollable variables as environmental variables as they are externally given within the ambient environment of the experiment. Examples of such uncontrollable variables are temperature, humidity and wind speed. This section presents an extension to Algorithm~\ref{alg:bo} that allows the inclusion of uncontrollable environmental variables in the optimisation process. The extension can be broken down into three parts as highlighted in Algorithm~\ref{alg:env-bo}.

The main modification to Algorithm~\ref{alg:bo} concerns the surrogate modelling. The basic Bayesian optimisation algorithm fits a surrogate model over all controllable variables. Environmental variables are not included and assumed to be fixed over the full optimisation process or irrelevant to the output. Algorithm~\ref{alg:env-bo} does not make this assumption and includes all controllable parameters $\boldsymbol x_C$ and environmental variables $\boldsymbol x_E$ in its surrogate model. The inputs $\boldsymbol X_n$ of the training data $\mathcal{D}_n = \{ \boldsymbol X_n, \boldsymbol y_n \}$ are extended from $\boldsymbol X_n = \{\boldsymbol X_{n, C}\}$ to $\boldsymbol X_n = \{\boldsymbol X_{n, E}, \boldsymbol X_{n, C}\}$.

The second extension regards the computation of the next candidate point, specifically, the maximisation of the acquisition function. While in Algorithm~\ref{alg:bo} all parameters are assumed to be controllable and the acquisition function can be maximised over all parameters $\max_{\boldsymbol x_{C}} \alpha (\boldsymbol x_{C})$, Algorithm~\ref{alg:env-bo} must differentiate between the controllable parameters $\boldsymbol x_C$ and environmental variables $\boldsymbol x_E$. The uncontrollable variables are given by the environment and can only be measured but not manipulated. Hence, the maximisation of the acquisition function is broken down into two steps. First, the environmental variables are measured. This gives values for the uncontrollable inputs for the next candidate point $\boldsymbol x_{n+1, E}$. Second, the acquisition function is maximised conditional on these values for the environmental inputs $\max_{\boldsymbol x_{C} \vert \boldsymbol x_{E}} \alpha (\boldsymbol x_{C})$ resulting in the controllable inputs for the next candidate point $\boldsymbol x_{n+1, C}$. Conditional maximisation essentially means that the environmental variables $\boldsymbol x_E$ are treated as fixed for the maximisation of the acquisition function for one iteration. This assumes that the environmental variables do not change significantly from the time of measuring until the evaluation of the new candidate point $\boldsymbol x_{n+1}$. This assumption should be realistic for most experiments, as one iteration of the Bayesian optimisation loop takes only a few seconds. However, issues could arise when working with environmental variables that change rapidly. The new candidate point $\boldsymbol x_{n+1}$ is then defined as a combination of the measurements for the environmental variables $\boldsymbol x_{n+1, E}$ and the results of the maximisation of the acquisition function $\boldsymbol x_{n+1, C}$.

The last adjustment to Algorithm~\ref{alg:bo} focuses on the generation of the training data. Usually, training data is produced using a space-filling design, such as a Latin hypercube \citep{McKay2000}. Under the assumption that all parameters can be controlled, the experiment can be repeated for each training data point to observe its output. The modified Bayesian optimisation algorithm, however, includes uncontrollable variables in its computation. Thus, it is not possible to evaluate any arbitrary combination of inputs as it is limited by the current measurement of the environmental variables. To resolve this issue, Algorithm~\ref{alg:env-bo} uses one training data point $\boldsymbol x_0$ instead of multiple points generated from a space-filling design. The initial training data is restricted to a single point, while the second data point is already computed via Bayesian optimisation. The first data point is generated by taking measurements for the environmental variables $\boldsymbol x_{0, E}$ and randomly selecting values for the controllable parameter $\boldsymbol x_{0, C}$. These inputs are then evaluated resulting in a complete training inputs-output pair $\mathcal{D}_0 = \{ \boldsymbol x_0, y_0 \}$, followed by the first Bayesian optimisation loop.

\begin{algorithm}
\caption{Modified Bayesian optimisation algorithm with environmental conditions}\label{alg:env-bo}
\begin{algorithmic}

\Require Evaluation budget $N$, surrogate model $\mathcal{M}$, acquisition function $\alpha$.

\State \hl{Sample an initial training data point $\boldsymbol x_0 = \{\boldsymbol x_{0, D}, \boldsymbol x_{0, C}\}$ where environmental parameters $\boldsymbol x_{0, E}$ are measured and controllable parameters $\boldsymbol x_{0, C}$ are randomly sampled} and gather observation $y_0$.

\State Set $n = 0$.

\While{$n \leq N -n_0$}

\State \hl{Fit surrogate model $\mathcal{M}$ to training data $\mathcal{D}_n = \{ \boldsymbol X_n, \boldsymbol y_n \}$, where $\boldsymbol X_n = \{\boldsymbol X_{n, E}, \boldsymbol X_{n, C}\}$.}
\State \hl{Measure environmental variables $\boldsymbol x_{n+1, E}$.}
\State \hl{Find values for the controllable parameters $\boldsymbol x_{n+1, C}$ that maximise the acquisition function $\alpha$ conditionally on the measurements $\boldsymbol x_{n+1, E}$ such that $\boldsymbol x_{n+1} = \{\boldsymbol x_{n+1, E}, \boldsymbol x_{n+1, C}\}$, i.e., solve $\max_{\boldsymbol x_C \vert \boldsymbol x_E} \alpha (\boldsymbol x_C)$.}
\State Evaluate $\boldsymbol x_{n+1}$ by observing $y_{n+1}$  
\State Increment $n$.

\EndWhile

\State \Return Point $\boldsymbol x^*$ with highest observation $y^*$.
\end{algorithmic}
\end{algorithm}

In contrast to \cite{Krause2011}, the proposed approach does not assume different covariance structures for controllable and environmental variables. When working with experiments and simulators, the underlying objective function is generally unknown or too complex to compute directly. Even with expert knowledge, there might not be enough information about these black boxes to confidently assume a linear or additive structure for the environmental variable \citep{Frazier2018}. Hence, it is important to provide the surrogate model with enough flexibility to estimate the covariance structure itself. This can be achieved by using the Mat\'ern kernel for controllable and environmental variables---or the radial basis function kernel for very smooth objective functions.

\begin{figure}
    \centering
    \includegraphics[width=\linewidth]{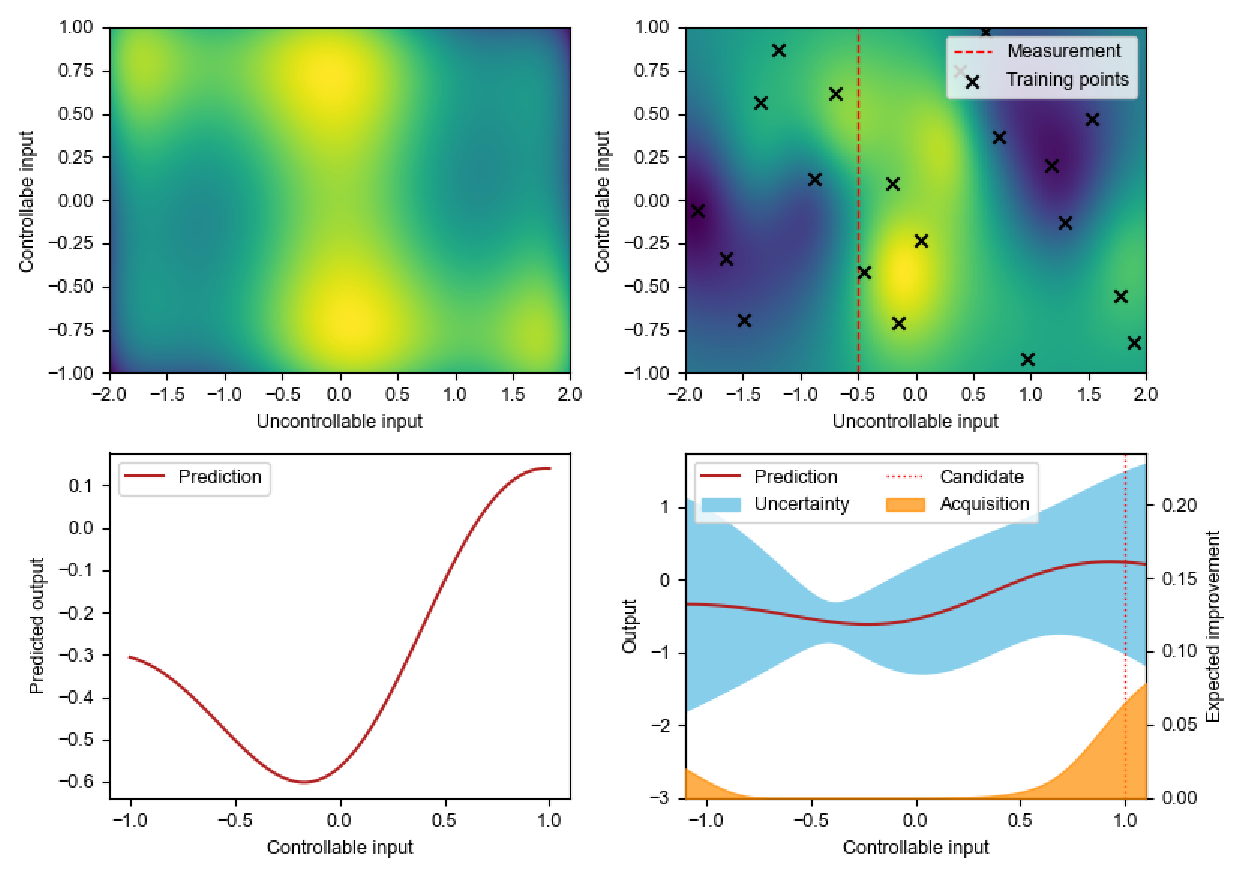}
    \caption{Maximisation of a two-dimensional problem with one environmental variable $x_1$ and one controllable variable $x_2$. Yellow areas indicate high outputs and dark blue areas indicate low outputs. Upper-left: True objective function. Upper-right: Prediction of a Gaussian process with a measurement taken for the next conditional optimisation step. Lower-left: Gaussian process prediction for optimisation conditional on the measurement. Lower-right: Bayesian optimisation step conditional on the measurement}
    \label{fig:env-bo}
\end{figure}

Figure~\ref{fig:env-bo} illustrates the conditional variable optimisation on a two-dimensional problem with one uncontrollable variable $x_1$ and one controllable parameter $x_2$. The upper-left plot shows the true output of the objective function, where yellow areas indicate high function values and blue areas indicate low function values. The goal is to find the optimal value for the controllable parameter (y-axis) that maximises the output for any value of the uncontrollable variable (x-axis). The upper-right plot shows the predictive mean of a Gaussian process fitted to $20$ training data points (black crosses). Following Algorithm~\ref{alg:env-bo}, a measurement (red dashed line) of the uncontrollable variable is taken resulting in $x_1 = -0.5$. The next iteration of the optimisation loop is performed conditional on this measurement. The lower-left plot shows the predictive mean of the Gaussian process for $x_1 = -0.5$. The conditional optimisation essentially takes a slice from the full surrogate model and reduces the two-dimensional optimisation problem to a one-dimensional problem for each iteration of the loop, where only the controllable parameters are considered. However, the information gained from the data is shared between each iteration. Notice that no training points lie on the measurement line but the model uses the available training points to inform its prediction. If Algorithm~\ref{alg:bo} were used, the uncontrollable input would be assumed to be fixed for the full optimisation loop and the optimisation process would need repeating for each value of $x_1$. The lower-right plot extends the lower-left plot by adding the uncertainty from the Gaussian process and the acquisition function. The optimal value of the controllable input $x_2$ is found by maximising the acquisition function and the new candidate point is a combination of this maximum and the measurement taken for the uncontrollable variable $x_1$. The candidate point is observed and added to the training data to be used in the next iteration of the optimisation loop.

\section{Simulations}\label{sec:simulations}
This section introduces two synthetic test functions\footnote{See \url{https://www.sfu.ca/~ssurjano/optimization.html} for further details on the synthetic test functions.}---the Levy function and the Hartmann function---and applies the Bayesian optimisation algorithm with environmental conditions presented in Section~\ref{sec:env-bo}. Simulations for both problems are run for $100$ evaluations and are repeated $30$ times\footnote{Runs, replications, repeats are used interchangeably in this article.} to validate the robustness of Algorithm~\ref{alg:env-bo}. This decreases the risk that results are influenced by the method's inherent randomness, e.g., the randomly sampled training points that initialise the algorithm. Both problems assume one uncontrollable variable whose value is provided by a random walk at each iteration. In the simulations, each step of the random walk adds a sample from a uniform distribution $\mathcal{U}$ to the previous value of the uncontrollable variable, such that
\begin{equation} \label{eq:uniform}
    \boldsymbol x_{n,E} = \boldsymbol x_{n-1,E} + \mathcal{U}_{[-\boldsymbol a, \boldsymbol a]},
\end{equation}
where $\boldsymbol a$ is a vector of small predefined constants that provide the minimal and maximal change of the environmental variables from one iteration to the next. This uniform assumption represents the natural fluctuation of the uncontrollable variables encountered in a real-world application, e.g., changes in temperature, humidity and wind speed. It further allows the investigation of uncontrollable variables with different fluctuation levels by increasing or decreasing the constants in  $\boldsymbol a$ as discussed in Section~\ref{sec:fluctuation}. Assuming another distribution for the constants in $\boldsymbol a$, such as a Normal distribution, is an alternative to this approach.

Subsequent sections analyse the performance of Algorithm~\ref{alg:env-bo} by comparing predictions from the Gaussian process models $\mu_n(\boldsymbol x)$ to the true optimal values $f(\boldsymbol x)$. In both cases, results are obtained by maximising the Gaussian process prediction and the true objective function conditional on identical test values $\boldsymbol x^\prime_E$ for the uncontrollable variable. These test values are sampled from a maximin Latin hypercube design \citep{McKay2000} within the observed domain of the uncontrollable variable $\left[ \min(\boldsymbol x_{N,E}), \max(\boldsymbol x_{N,E}) \right]$ considered by the algorithm. We also call this observed domain the effective domain. This ensures a fair comparison by avoiding predictions outside of the effective domain as these require extrapolation. Extrapolation with Gaussian processes generally means that predictions default to the prior mean function. In cases where extrapolation cannot be avoided, making the prior mean function as informative as possible---for example, by going beyond zero and constant mean functions with polynomial and trigonometric mean functions---can improve results significantly \citep{Planas2020}. To score the performance of the algorithm, the mean absolute percentage error $\text{MAPE}(\mu_n(\boldsymbol X^\prime_i), f(\boldsymbol X^\prime_i)) = \frac{1}{m} \sum_{i=1}^m \lvert \frac{(\mu_n(\boldsymbol x^\prime_i) - f(\boldsymbol x^\prime_i)}{f(\boldsymbol x^\prime_i)} \rvert$ between the Gaussian process' predictions and the truths are computed for all $m$ test points $\boldsymbol X^\prime$.

The acquisition criterion conditional on values of the uncontrollable variables $\boldsymbol x_{n,E}$ is maximised with the SLSQP algorithm \citep{Kraft1994} using multiple starts. For this strategy, $100$ points are sampled from a maximin Latin hypercube design \citep{McKay2000} and evaluated by the acquisition criterion. The best $20$ points are then used to initialise the SLSQP algorithm and only the best result is used as the solution for the optimisation problem. The multiple starts aim to reduce the risk of converging towards a local optimum instead of the desired global optimum of the acquisition function.

\subsection{The two-dimensional Levy function}\label{sec:camel}
\begin{figure}
    \centering
    \includegraphics[width=0.8\linewidth]{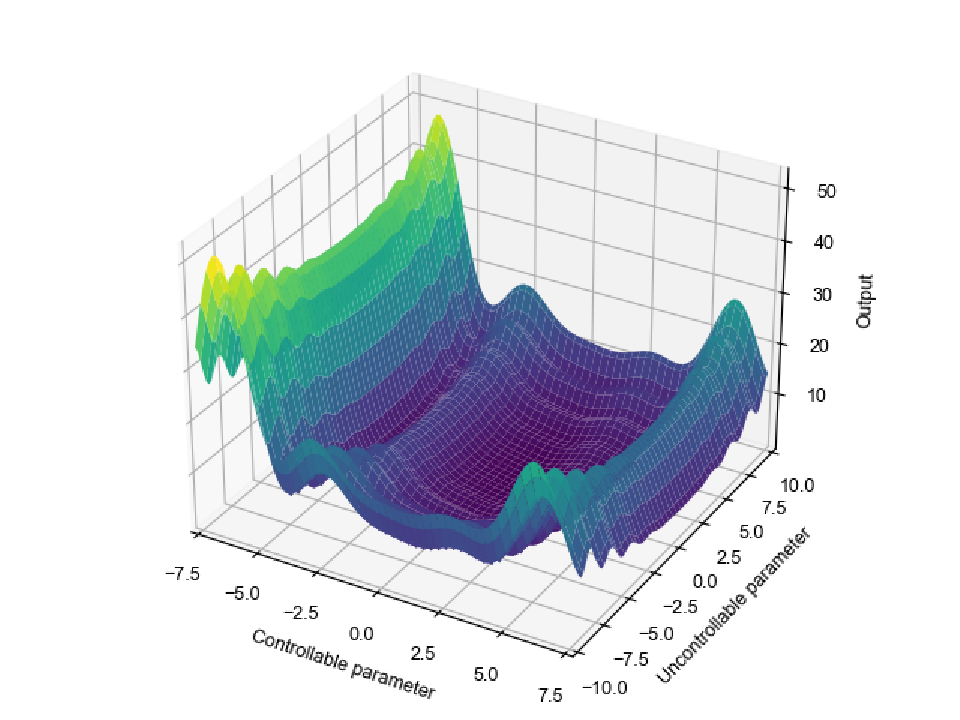}
    \caption{Two-dimensional negated Levy function with one controllable parameter $x_1$ bounded by $[-7.5, 7.5]$ and one uncontrollable variable $x_2$ bounded by $[-10, 10]$}
    \label{fig:levy}
\end{figure}

The Levy function
$$f(\boldsymbol x) = \sin^2 (\pi w_1) + (w_1 - 1)^2 [1 + 10 \sin^2 (\pi w_1 + 1)] + (w_2 - 1)^2 [1 + \sin^2(2\pi w_2)],$$
where $w_i = 1 + \frac{x_i - 1}{4}$, for $i = 1, 2$, is a two-dimensional function with two input parameters $\boldsymbol x_1$ and $\boldsymbol x_2$. Although often used as a minimisation problem to find the global minimum $f(\boldsymbol{x}^*) = 0$ at $\boldsymbol{x}_1^* = (1, 1)$ for the input space $[-10, 10]^2$, we choose the bounds of the parameters as $[-7.5, 7.5]$ and $[-10, 10]$ respectively to create a maximisation problem that is better suited for testing Algorithm~\ref{alg:env-bo}. The function given in Figure~\ref{fig:levy} shows a clear ridge at values of about $-6$ for the controllable parameter $x_2$ for all values of the uncontrollable variable $x_1$. The simulations use $a = 1.5$ as the uniform distribution constant of the random walk for the uncontrollable variable $x_2$. The effect of setting $a$ to different values is discussed in Section~\ref{sec:fluctuation}.

\begin{figure}
    \centering
    \includegraphics[width=\linewidth]{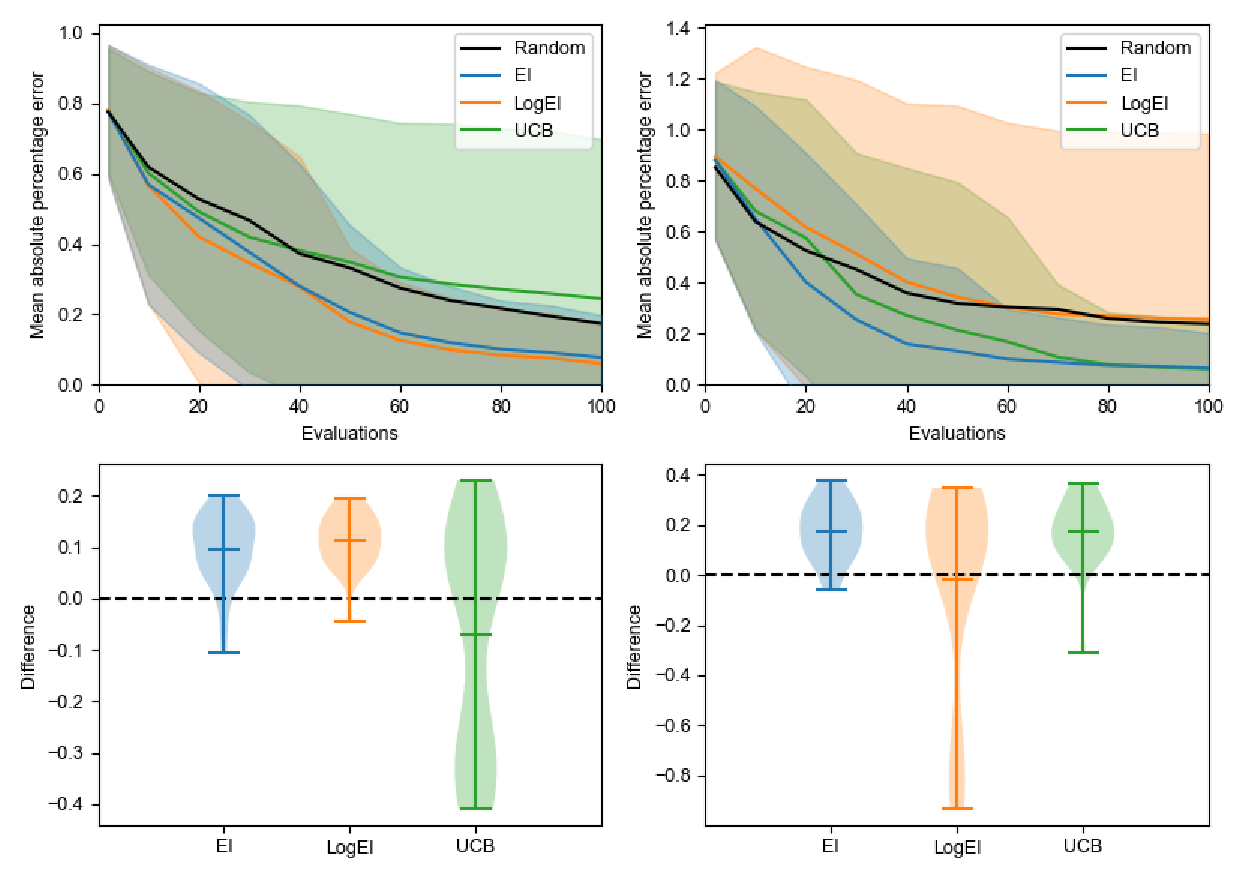}  
    \caption{Upper row: Means (lines) and $95$\% confidence intervals (shaded areas) of the mean absolute percentage error between Gaussian process prediction and truth over $30$ replications. Lower row: Difference between algorithms and random benchmark after $100$ function evaluations for each of the $30$ replications. Two-dimensional Levy function with one uncontrollable parameter on the left and six-dimensional Hartmann function with one uncontrollable parameter on the right}
    \label{fig:results}
\end{figure}

\begin{figure}
    \centering
    \includegraphics[width=\linewidth]{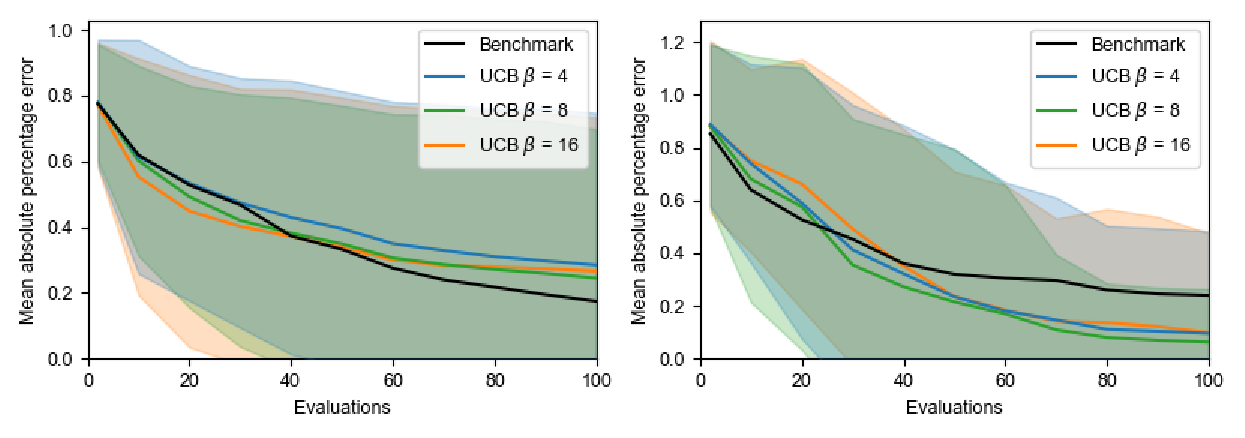}  
    \caption{Comparison of different trade-off parameters $\beta$ for the upper confidence bound acquisition function. Means (lines) and $95$\% confidence intervals (shaded areas) of the mean absolute percentage error between Gaussian process prediction and truth over $30$ replications. Two-dimensional Levy function with one uncontrollable parameter on the left and six-dimensional Hartmann function with one uncontrollable parameter on the right}
    \label{fig:ucb}
\end{figure}

The upper-left plot in Figure~\ref{fig:results} shows the performance of Algorithm~\ref{alg:env-bo} as the mean absolute percentage error between the maximum of the Gaussian process prediction and the maximum of the true objective function conditional on $25$ test values of the uncontrollable variable. Three alternatives for the acquisition function---expected improvement (EI), log expected improvement (LogEI) and upper confidence bound with a trade-off parameter $\beta=8$ (UCB) introduced in Section \ref{sec:bo-acq}---are compared to a benchmark where values for the controllable variable were selected randomly. The solid lines indicate the mean performance while the shaded areas indicate the 95\% confidence interval over the $30$ replications. For each replication, the mean absolute percentage error is computed for every $10$ evaluations for the same $25$ test points $\boldsymbol X^\prime_E$ of the environmental variable. The test values are sampled from a Latin hypercube bounded by the minimal and maximal value of the uncontrollable variable after $100$ function evaluations. The mean absolute percentage error starts just below $0.8$ for all alternatives. While the improvement-based algorithms (EI and LogEI) performed better than the random benchmark, the algorithm using upper confidence bound performs worse. After $100$ function evaluations, expected improvement and log expected improvement have a mean absolute percentage error of $0.08$ and $0.06$ respectively and improve upon the random benchmark ($0.17$). Upper confidence bound only achieves a mean absolute percentage error of $0.24$. Moreover, the optimistic strategy shows large confidence intervals indicating that the method is not robust. Altering the trade-off parameter $\beta$ did not improve this result as illustrated in the left plot in Figure~\ref{fig:ucb}.

The lower-left plot presents the difference between the mean absolute percentage error of the random benchmark and the three alternative acquisition functions after $100$ evaluations. This difference is computed for each of the $30$ replications and distributions of the differences are plotted for the three different acquisition functions.  Negative values indicate replications where the benchmark resulted in superior solutions, while positive values indicate that the given version of Algorithm~\ref{alg:env-bo} performed better than the benchmark.  Although no alternative is better than the benchmark for every single replication of the $30$ total replications, there is a clear difference between the improvement-based and the optimistic acquisition functions. Indeed the mean of upper confidence bound (displayed by the horizontal line towards the center of each violin plot) is the only one worse than $0$. This means that the random benchmark outperforms upper confidence bound on average. Additionally, the plot mirrors the lack of robustness discovered earlier by the large spread in differences. While the method performs better for some replications than the random benchmark, it performs much worse for others. A Mann-Whitney U test was performed to determine what alternatives perform differently from the random benchmark to a $1$\% significance level. The test returned a $p$-value of $0.0$ for the improvement-based algorithms and $0.37$ for the upper confidence bound. This rejects the hypothesis that methods perform equally well for the improvement-based methods reinforcing the results from the visual analysis that expected improvement and log expected improvement perform significantly better than the random benchmark. This cannot be said for the optimistic method---the test cannot reject the null hypothesis indicating that upper confidence bound does not perform significantly differently from a random approach.

\subsection{The six-dimensional Hartmann function}\label{sec:hartmann}
The negated\footnote{Bayesian optimisation problems are generally expressed as maximisation problems as introduced in Section \ref{sec:methodology}. The Hartmann function is a minimisation problem and is negated for the following simulations to keep in line with this convention.} Hartmann function
$$f(\boldsymbol{x}) = \sum^4_{i=1} \alpha_i \exp \left(- \sum^6_{j=1} \boldsymbol{A}_{ij} \left(x_j - \boldsymbol{P}_{ij}\right)^2 \right),$$ \label{eq:Hartmann}
where
\begin{align*} 
    \alpha = &(1.0, 1.2, 3.0, 3.2)^T,\\
    \boldsymbol{A} = &\begin{pmatrix}
                        10.00 &  3.00 & 17.00 &  3.50 &  1.70 &  8.00\\
                         0.05 & 10.00 & 17.00 &  0.10 &  8.00 & 14.00\\
                         3.00 &  3.50 &  1.70 & 10.00 & 17.00 &  8.00\\
                        17.00 &  8.00 &  0.05 & 10.00 &  0.10 & 14.00
                     \end{pmatrix}, \text{ and}\\
    \boldsymbol{P} = 10^{-4} &\begin{pmatrix}
                                1312 & 1696 & 5569 &  124 & 8283 & 5886\\
                                2329 & 4135 & 8307 & 3736 & 1004 & 9991\\
                                2348 & 1451 & 3522 & 2883 & 3047 & 6650\\
                                4047 & 8828 & 8732 & 5743 & 1091 &  381
                             \end{pmatrix},
\end{align*}
is a six-dimensional function with six input parameters $x_1$, $x_2$, $x_3$, $x_4$, $x_5$ and $x_6$ that are evaluated on the hypercube $(0, 1)^6$. It has six local maxima and one global maximum with $f(\boldsymbol{x}^*) = 3.32$ at $\boldsymbol{x}^* = (0.20, 0.15, 0.48, 0.28, 0.31, 0.66)$. The simulations use $a = 0.05$ as the uniform distribution constant of the random walk for the environmental variable $x_6$. Section~\ref{sec:fluctuation} considers different levels of $a$.

The upper-right plot in Figure~\ref{fig:results} shows the performance of Algorithm~\ref{alg:env-bo} over the $30$ repeats for the same three acquisition function as for the Levy function and compares it to the random benchmark. The mean absolute percentage error starts between $0.85$ and $0.90$ for all four algorithms. After $100$ function evaluations, the mean absolute percentage error between the prediction of the Gaussian process and the true objective value for Algorithm~\ref{alg:env-bo} using expected improvement and upper confidence bound with $\beta=8$ are $0.07$ and $0.06$ respectively---much better than the random benchmark with $0.24$. However, the algorithm using log expected improvement with a mean absolute percentage error of $0.26$ performs comparably to the benchmark on average. The large $95$\% confidence intervals for log expected improvement indicate that Algorithm~\ref{alg:env-bo} with log expected improvement is not robust in this case making it less reliable than expected improvement and upper confidence bound.

The violin plots of the difference between the mean absolute percentage error of the benchmark and the three variations of Algorithm~\ref{alg:env-bo} for all $30$ replications on the lower-right in Figure~\ref{fig:results} indicate that expected improvement performs the best with almost all replications better than the random benchmark. Log expected improvement on the other hand performs comparably to the benchmark on average but has a large spread with some replication performing much worse. Upper confidence bound performs similarly to expected improvement but has an outlier that performs much worse than the random benchmark. Overall, expected improvement presents itself as the most robust method that is not prone to outliers. Despite these differences between algorithms, results after $100$ evaluations for all three algorithms are significantly different from the random results to a $1\%$ significance level: the $p$-values from Mann-Whitney U tests are $0.0$ for the expected improvement and the upper confidence bound version of Algorithm~\ref{alg:env-bo} and $0.07$ log expected improvement. While this shows that the results of all methods are significantly different from the benchmark, only expected improvement and upper confidence bound perform better than the benchmark indicated by the better average mean absolute percentage error.

Figure~\ref{fig:ucb} shows results for different values of the trade-off parameter $\beta$. While the average performance is very similar after $100$ evaluations, there is a difference in the $95$\% confidence intervals for the Hartmann function. However, no clear correlation is noticeable as $\beta=8$ performs better than $\beta=4$ and $\beta=16$ indicating no clear trend.

\section{Empirical analysis of properties}\label{sec:empirical}
Based on the investigations of Section~\ref{sec:empirical}, we define ENVBO as a version of Algorithm~\ref{alg:env-bo} that uses the expected improvement acquisition function. This section explores the sensitivity to five properties of ENVBO using the negated Hartmann function. Firstly, Section~\ref{sec:noise} analyses the effect of adding different levels of random Gaussian noise to the function's output. Secondly, Section~\ref{sec:multi} investigates the influence of having more than one uncontrollable variable. Thirdly, Section~\ref{sec:fluctuation} examines the significance of the fluctuation level, i.e. the step size $a$ of the random walk. Fourthly, Section~\ref{sec:variability} studies the impact of the variability in the uncontrollable variable. Lastly, Section~\ref{sec:domain} considers the relationship between the algorithm performance and the effective domain size of the uncontrollable variables. An off-the-shelf version of ENVBO is available via the open-source Python package NUBO \citep{Diessner2023}.

To make the comparison fair, the same $30$ initial starting points and random walks were used for the $30$ runs. For the random walks, this is achieved by sampling changes in percentages and scaling them by the maximal step size $a$, rather than sampling absolute values directly. 

\begin{figure}
    \centering
    \includegraphics[width=\linewidth]{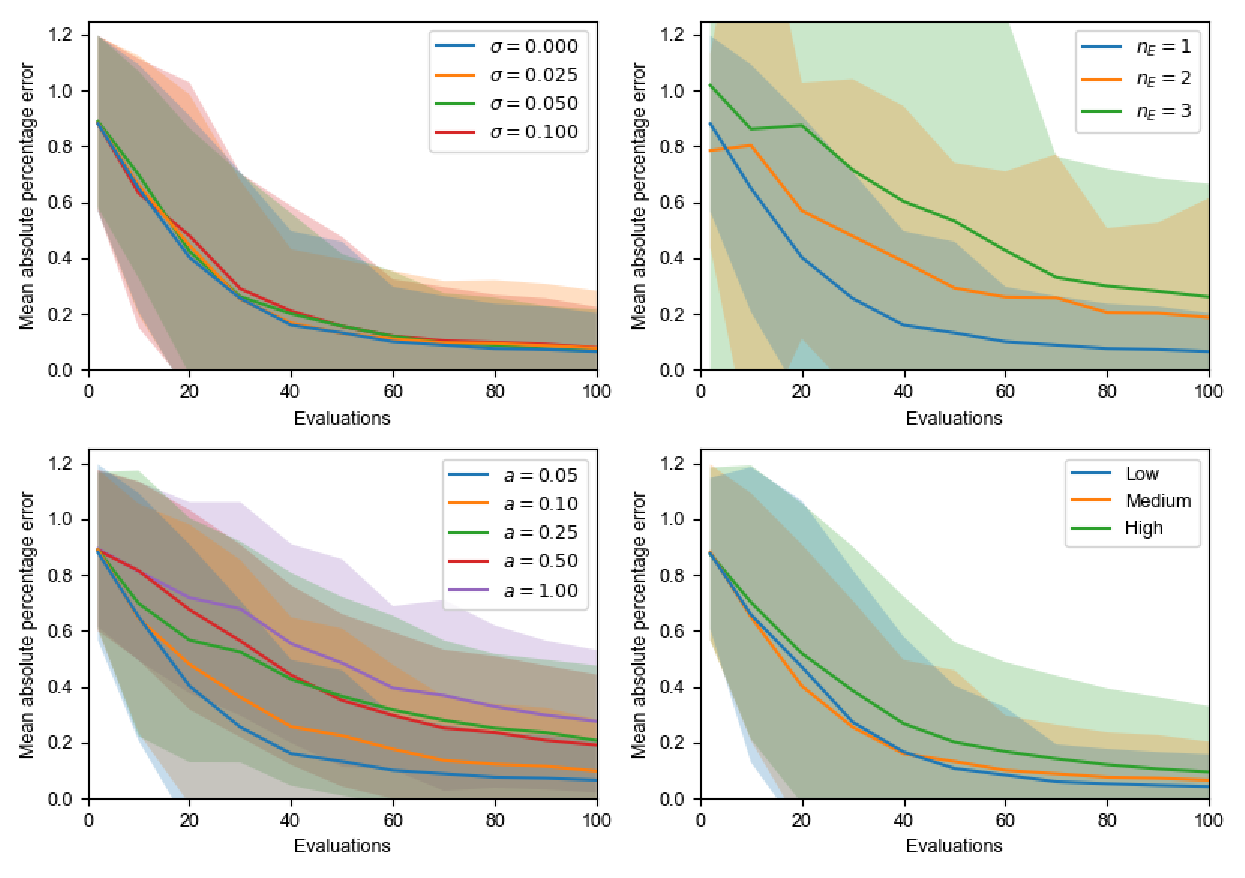}
    \caption{Means (lines) and $95$\% confidence intervals (shaded areas) of the mean absolute percentage error between the predictive mean of the Gaussian process and the truth over $30$ replications for the six-dimensional Hartmann function. Upper-left: Comparison of randomly added noise levels, $\mathcal{N} \left(0, \sigma^2 \right)$. Upper-right: Comparison of different numbers of uncontrollable parameters $n_E$. Lower-left: Comparison of five different step sizes $a$ for the random walk $\mathcal{U}_{[-a, a]}$ added to the previous uncontrollable value. Lower-right: Comparison of uncontrollable variables with different parameter variability}
    \label{fig:analysis-1}
\end{figure}

\subsection{Noise}\label{sec:noise}
Most physical experiments in engineering cannot be conducted without introducing some noise, such as measurement uncertainty, that cannot be eliminated entirely. This section replicates this situation by adding some randomly generated noise $\epsilon$ to the Hartmann function. The noisy function can be defined as $g(\boldsymbol x) = f(\boldsymbol x) + \epsilon$, where $f(\boldsymbol x)$ is a deterministic negated Hartmann function from Equation~(\ref{eq:Hartmann}). The noise is sampled from a Normal distribution centred around $0$ with a small standard deviation $\sigma$, such that $\epsilon \sim \mathcal{N}(0, \sigma^2)$. The simulations explore noise levels with $\sigma=0.00$, $\sigma=0.025$, $\sigma=0.050$, and $\sigma=0.100$. Considering the range of the Hartmann function, this corresponds to standard deviations of $1.5\%$, $3.0\%$ and $6.0\%$ of the full output range respectively. This means that for any of these three cases, $68.3\%$ of the added noise values will fall between $\pm 1\sigma$, $95.5\%$ fall between $\pm2\sigma$ and $99.7\%$ fall between $\pm 3\sigma$. For $\sigma=0.100$, this translates into noise values that decrease or increase the real output by up to $6.0\%$ of the output range $68.3\%$ of the time, by $12.0\%$ of the output range $95.5\%$ of the time, and by $18.1\%$ of the output range $99.7\%$ of the time.

The plot in the upper-left of Figure~\ref{fig:analysis-1} shows the performance of ENVBO for each of these four noise levels. The results indicate no significant difference in the average performance or the $95\%$ confidence intervals between the four cases. Overall, the proposed method does not seem to be sensitive to adding modest noise levels.

\subsection{Number of uncontrollable variables}\label{sec:multi}
For some experiments, there might be more than one influential uncontrollable variable present. The upper-right plot of Figure~\ref{fig:analysis-1} provides results about the performance of ENVBO with one, two and three uncontrollable variables while the overall dimensionality of the problem stays the same at $n = 6$. For $n_E = 1$, input six of the Hartmann function from Equation~(\ref{eq:Hartmann}) is assumed uncontrollable, while input one is added to the uncontrollable variables for $n_E = 2$, and input one and four are added for $n_E = 3$. The number of test points is increased from $25$ to $50$ for $n_E = 2$ and to $75$ for $n_E = 3$. The results show that the mean absolute percentage error increases with increasing numbers of environmental variables. Particularly, the $95\%$ confidence intervals widen significantly. This result is expected as the input space of the environmental variables grows exponentially with $n_E$ and requires exponentially more training points to cover the input space equally well as lower $n_E$. Thus, more evaluations are required to achieve similar results.

In higher dimensional space, there exists also the problem of extrapolation with Gaussian processes. Test points are generated with a Lain hypercube that uses minimal and maximal values of all environmental variables as their upper and lower bounds. With increasing numbers of the environmental variables $n_E$, it is very likely that while values for individual dimensions fall within these bounds, the combination of values for different dimensions falls in areas that were not explored by the optimisation algorithm. To predict outputs for these test points, the final Gaussian process model will extrapolate. As mentioned in Section~\ref{sec:simulations}, extrapolation with Gaussian processes generally means that predictions default to the prior mean function. 

\subsection{Fluctuation}\label{sec:fluctuation}
Uncontrollable variables will fluctuate to different extents from one evaluation to the next, for example, when measured in physical experiments. Higher fluctuations cause big jumps in the uncontrollable variable values, while uncontrollable variables will be more stable for lower fluctuations. The lower-left plot of Figure~\ref{fig:analysis-1} shows results for five different fluctuation levels implemented by varying parameter $a$ of the uniform distribution in equation~(\ref{eq:uniform})---the higher $a$ the higher the fluctuation of the uncontrollable variable. For higher fluctuations, there is more potential for larger effective domains of the uncontrollable variable, i.e., the value range for which the environmental variables are explored. The results show that Algorithm~\ref{alg:env-bo} performs better for lower $a$ and the performance of the final Gaussian process models decreases with increasing fluctuation. This can be explained by considering the effective domain. For $a = 1.00$, any value of the domain can be randomly selected at each iteration and the whole domain will probably be searched. For $a = 0.05$, only values that differ by $0.05$ from the previous evaluation can be randomly selected making it less probable that the whole domain will be searched before the evaluation budget is exhausted. If a Gaussian process is fitted to domains with different sizes but the number of data points (evaluations) remains the same it is probable that predictions will be better for smaller domains. For example, it is plausible that a Gaussian process fitted to $10$ data points within the domain $[0, 0.1]$ will reflect the truth in this domain better than a Gaussian process fitted to $10$ data points within the much larger domain $[0, 1]$ assuming that everything else stays comparable. Thus to improve the performance of high-fluctuating uncontrollable variables, more evaluations are required to match the performance of their low-fluctuating counterparts. 

\subsection{Parameter variability}\label{sec:variability}
Parameter variability is closely linked to fluctuation and indicates how quickly the parameter value changes when moving along the axis. Uncontrollable variables with a low variability will only change very slightly, while variables with a high variability will change considerably. Value changes of two variables could differ considerably for the same fluctuation level when they have different levels of parameter variability. As a proxy for the parameter variability, the length-scales of a well-fitting Gaussian process over the full domain are considered. The length-scales quantify how long a certain parameter is correlated when moving along its axis \citep{Gramacy2020}. Consider, for example, a parameter with a large length-scale. When this parameter value is changed by a certain amount, a relatively small change is expected in the output---provided everything else stays the same. For a parameter with a large length-scale, the change in the output is expected to be larger. To achieve a well-fitting model, a Gaussian process is fitted to $2000$ data points sampled from a Latin hypercube \citep{McKay2000}. The resulting length-scales for all six parameters are in order $1.30$, $1.90$, $4.81$, $1.48$, $1.51$, and $1.47$. The first input has the lowest length-scale suggesting that values are only correlated for a short distance when moving along its axis. This means that the parameter variability of the first input is high. In contrast, the third input has the highest length-scale indicating a low parameter variability. Moving along its axis less change is expected for the third input than for the first input.

The lower-right plot of Figure~\ref{fig:analysis-1} compares the first (high), third (low) and sixth input (medium) when chosen as the uncontrollable variable. Differences in the low and medium parameter variability are very small, while there is some difference compared to the parameter with a high variability: the confidence interval is noticeably larger and the average of the mean absolute percentage error is slightly worse, especially for evaluations $30$ to $80$.

\subsection{Domain size}\label{sec:domain}
\begin{figure}
    \centering
    \includegraphics[width=\linewidth]{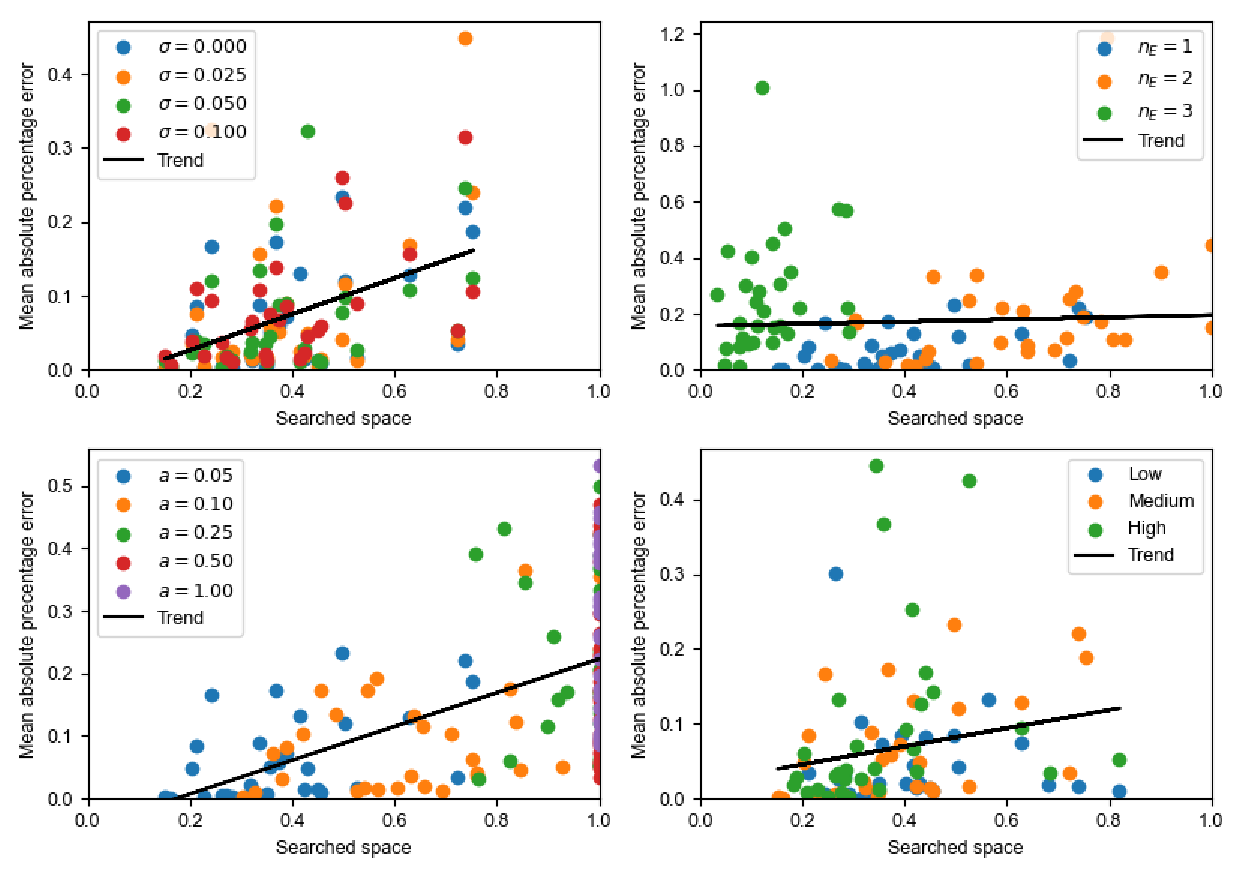}
    \caption{Relationships between the actual effective domain of the uncontrollable variables and the mean absolute percentage error of an individual run for the six-dimensional Hartmann function. Upper-left: Comparison of randomly added noise levels, $\mathcal{N} \left(0, \sigma^2 \right)$. Upper-right: Comparison of different numbers of uncontrollable parameters $n_E$. Lower-left: Comparison of five different step sizes $a$ for the random walk $\mathcal{U}_{[-a, a]}$ added to the previous uncontrollable value. Lower-right: Comparison of uncontrollable variables with different parameter variability}
    \label{fig:analysis-2}
\end{figure}

This section investigates the relationship between the size of the effective domain, that is the actual searched input space of the environmental variables, and the performance of the Gaussian process as touched on in Section~\ref{sec:fluctuation}. Figure~\ref{fig:analysis-2} uses the same data as Figure~\ref{fig:analysis-1} but plots the effective parameter domain against the mean absolute percentage error which provides a proxy for the performance of the Gaussian process. Each point reflects one individual replication of the $30$ performed replications. The trend lines in each plot show that there is a positive relationship between the effective domain and the mean absolute percentage error. This means small effective domains generally correspond to small mean absolute percentage errors, while large effective domains generally correspond to large mean absolute percentage errors. Intuitively, this result makes sense as a Gaussian process should have a better fit for a smaller space than a larger space when the number of data points and the function in question stay the same. 

The results depicted in the lower-left plot reinforce the reasoning from Section~\ref{sec:fluctuation}---the larger the bounds of the uniform distribution $a$ that specify the maximal size of the random steps, the more potential for larger searched spaces. Thus, it is plausible that ENVBO performs better on smaller $a$ as the results of the lower-left plot of Figure~\ref{fig:analysis-1} suggest. 

Only the upper-right plot that plots the effective domain for different numbers of uncontrollable parameters against the mean absolute percentage error shows almost no relationship. Generally, the size of the environmental variable space grows exponentially with the number of environmental variables. This means that to achieve the same coverage for $n_E = 3$ as for $n_E = 1$ exponentially more evaluations are required. However, in this study, the evaluation budget is fixed to $100$ evaluations regardless of the number of environmental variables. Considering this, an inverse effect of the number of environmental variables and the effective domain would be expected. Yet, searched spaces for $n_E = 2$ are generally higher than for $n_E = 1$ and a rough order of $n_E = 2$, $n_E = 1$ and $n_E = 3$ for growing effective domains is noticeable. A possible reason is that the first environmental variable uses a fluctuation parameter $a = 0.05$ while all additional variables use $a = 0.1$. This affects the searched space significantly as depicted in the lower-left plot of Figure~\ref{fig:analysis-2} as discussed previously.

\section{Application to a wind farm simulator}\label{sec:application}
The power generation of wind farms is highly dependent on the wind speed, the wind direction and the placement of the individual wind turbines within a certain area or site \citep{Grady2005, Qureshi2023}. Most of the time, interest lies in finding the optimal wind turbine positions that maximise the energy production conditional on either constant or variable wind speeds and directions \citep{Mosetti1994, Chen2013, Parada2017}. Optimisation algorithms such as TOPFARM\footnote{See \url{https://topfarm.pages.windenergy.dtu.dk/TopFarm2/index.html} for further details on TOPFARM.} can be used when function evaluations are cheap, while methods mentioned in the introduction \citep{Willams2000, Swersky2013, ToscanoPalmerin2018} present a cost-effective alternative when function evaluations are expensive. However, these methods cannot find multiple solutions for different environmental conditions within one optimisation run which is the objective of this section. Specifically, this application aims to find the optimal wind turbine positions conditional on randomly changing wind directions resulting in one solution for each possible wind direction. While the wind speed is assumed to be fixed in this application, it could also be randomly changing. The result would be solutions for all combinations of wind direction and wind speed.

\begin{figure}
    \centering
    \includegraphics[width=\linewidth]{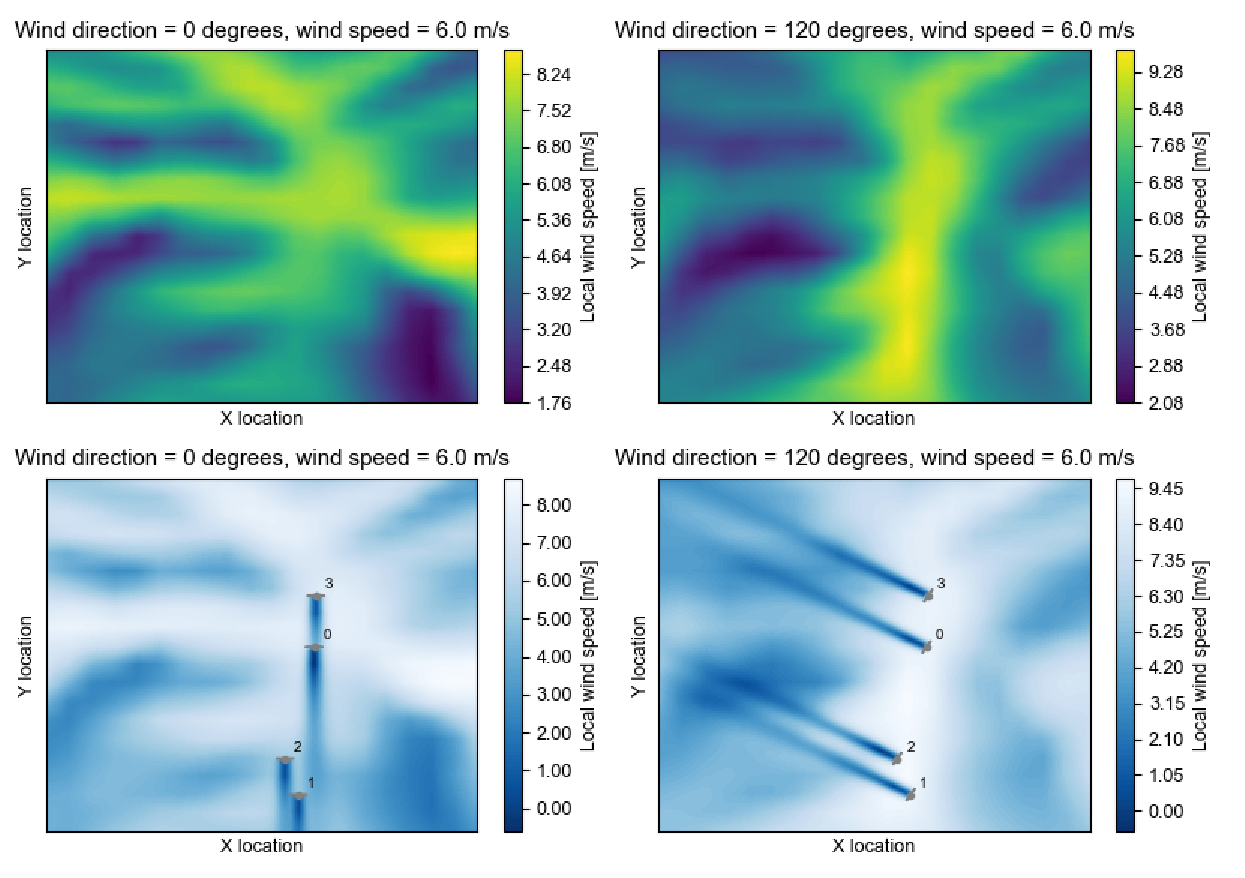}
    \caption{Wind farm simulator. Upper row: Local wind speed over the complex terrain for a wind direction of 0 and 120 degrees. Lower row: Wake of four wind turbines for a wind direction of 0 and 120 degrees. Wind speed is fixed at 6 m/s}
    \label{fig:simulator-1}
\end{figure}

The first row of plots of Figure~\ref{fig:simulator-1} shows the effect of the wind direction on the local wind speed for a complex underlying terrain while the global wind speed is fixed at $6$ m/s. The locations of the high local wind speeds necessary for high energy production shift significantly between a wind direction of $0$ and $120$ degrees. While there is a band of high local wind speeds down the centre of the X location for the latter, it roughly rotates $90$ degrees for the former. The plots show that ideal wind turbine positions are likely to differ for the two wind directions. Another important factor for maximising energy production is taking the wake of the wind turbines into account (lower row of Figure~\ref{fig:simulator-1}). While the wake of the wind turbines located upstream does not affect wind turbines located downstream for a wind direction of $120$, the wake of wind turbine $3$ for a wind direction of $0$ degrees heavily affects wind turbine $1$. This shows that although a wind farm position can be ideal for one wind speed, it can be suboptimal for another wind speed.

In this section, a fictitious site is considered with complex terrain as shown in Figure~\ref{fig:simulator-1} on which four wind turbines have to be placed to maximise the annual energy production (AEP). The wind direction is an environmental variable that varies according to a random walk as defined in Equation~(\ref{eq:uniform}) where $90$ and $135$ degrees are the lower and upper bounds respectively and wind directions can change by $\pm 5$ degrees from one iteration to the next. The global wind speed is fixed at $6$ m/s. In total, this results in a $9$-dimensional problem with eight controllable parameters---one X and one Y location for each of the four wind turbines---and the wind direction as the only environmental variable. Additionally, the positioning of the wind turbines is constrained such that wind turbines have to be at least $160$ metres apart to prevent the blades from colliding. Simulations are performed with PyWake \citep{PyWake2023} and use Vestas V80 wind turbines that can produce $2$MW of energy and have a blade diameter of $80$ metres.

ENVBO---a version of Algorithm~\ref{alg:env-bo} with expected improvement as its acquisition function---is run for a function evaluation budget of $200$ and benchmarked against regular Bayesian optimisation (Algorithm~\ref{alg:bo}) and the SLSQP optimisation algorithm \citep{Kraft1994}. Regular Bayesian optimisation, referred to as BO in the following paragraphs, uses expected improvement as its acquisition function as well to make comparisons fair. While ENVBO is capable of returning one solution for each wind direction after one optimisation run, BO and the SLSQP algorithm have to be run for each possible wind direction. To compare the algorithms, four wind directions are chosen---$90$, $105$, $120$ and $135$ degrees---and BO is restricted to $50$ function evaluations each to reach the same function evaluation budget as ENVBO. The function evaluations of the SLSQP algorithm cannot be restricted and the algorithm is therefore run until convergence. As BO and SLSQP are run for fixed wind directions, they do not make use of the random walk in contrast to ENVBO. This makes it easier for them to converge towards a solution as all influential variables can be controlled by the algorithms. ENVBO and BO were implemented via the open-source package NUBO \citep{Diessner2023} and an off-the-shelf version of ENVBO is available at \url{www.nubopy.com}. The SLSQP algorithm was implemented via the SciPy package \citep{SciPy2020}. Furthermore, all code for the optimisation of the wind farm simulator is available at \url{https://github.com/mikediessner/environmental-conditions-BO}.

\begin{figure}
    \centering
    \includegraphics[width=\linewidth]{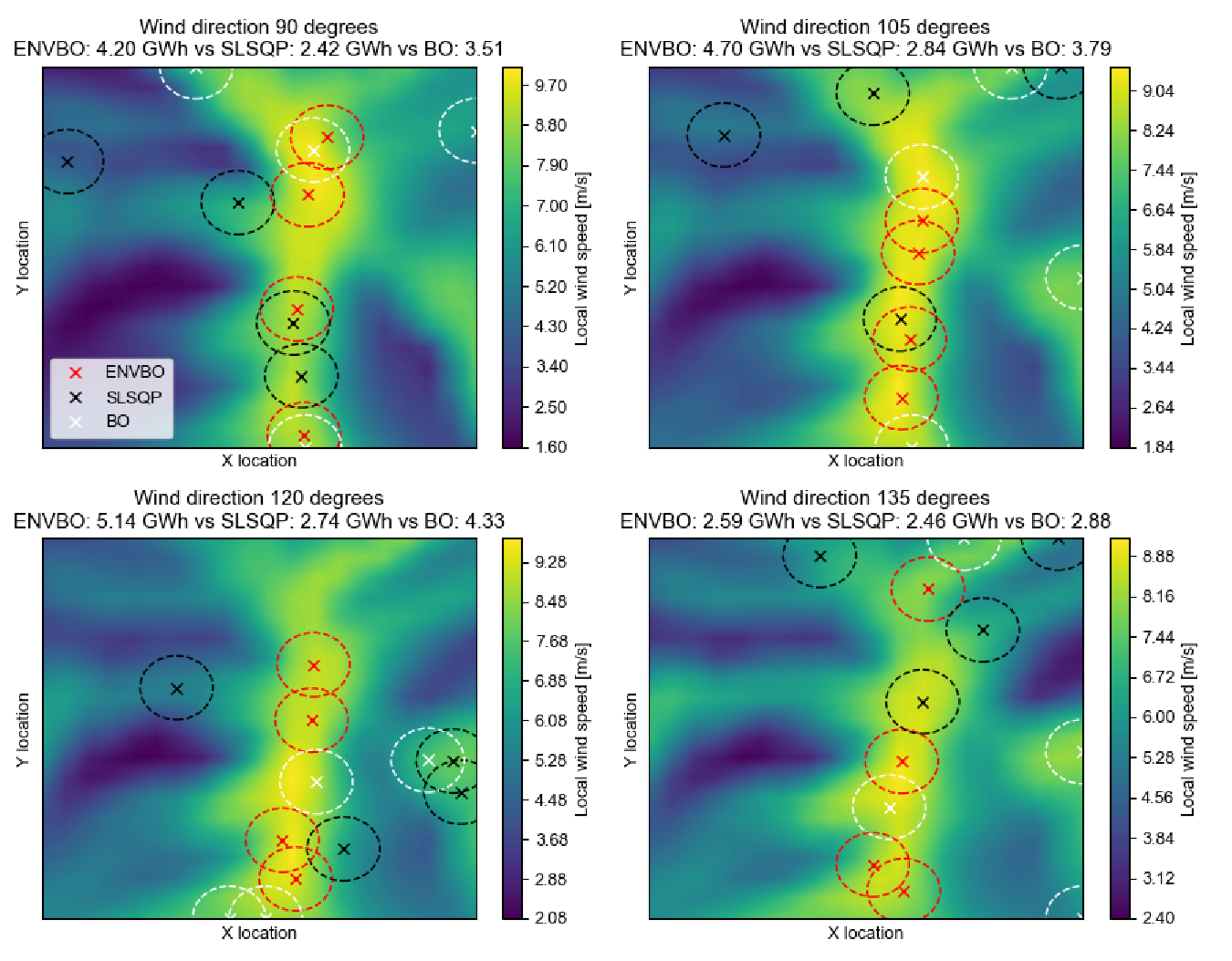}
    \caption{Annual energy production and placement of four wind turbines with spacing constraints. ENVBO (Algorithm~\ref{alg:env-bo}) is benchmarked against SLSQP and BO (Algorithm~\ref{alg:bo})}
    \label{fig:simulator-2}
\end{figure}

Figure~\ref{fig:simulator-2} shows the results for all three algorithms and all four wind directions. The dashed circles around the wind turbine positions indicated with crosses represent the placement constraint---no wind turbine of one colour can be placed within the dashed circle of another wind turbine with the same colour. For a wind direction of $90$ degrees, ENVBO performs best with an annual power production of $4.20$ GWh followed by BO and SLSQP with $3.51$ and $2.42$ GWh respectively. This is a $20\%$ improvement over BO and a $74\%$ improvement over SLSQP. SLSQP performs much worse than ENVBO and places at least one turbine in a subpar area with low local wind speeds, possibly due to converging towards a local maximum. While BO places three turbines in areas with high local wind speeds, it places one turbine within a suboptimal area. For a wind direction of $105$ degrees, ENVBO with $4.70$ GWh performs significantly better than BO and SLSQP which achieve $0.91$ and $1.86$ GWh less power generation respectively---a $24\%$ and $65\%$ improvement. A similar result is achieved for a wind direction of $120$ degrees. ENVBO outperforms BO and SLSQP by $0.81$ and $2.4$ GWh or $19\%$ and $88\%$ respectively. The results of the three strategies are closest for a wind direction of $135$ degrees. This is the only instance where ENVBO is beaten by another algorithm---in this case BO with an AEP of $2.88$ GWh. ENVBO achieves $0.29$ GWh ($10\%$) less AEP but still outperforms SLSQP by $0.13$ GWh ($5\%$). However, the differences are much smaller than for any of the first three wind directions. Considering Figure~\ref{fig:simulator-2}, ENVBO is the only strategy that consistently places all four wind turbines in the band of high local wind speeds that is located down the centre of the X location. For all other methods, at least one turbine is placed off to one side of this band.

\begin{figure}
    \centering
    \includegraphics[width=0.8\linewidth]{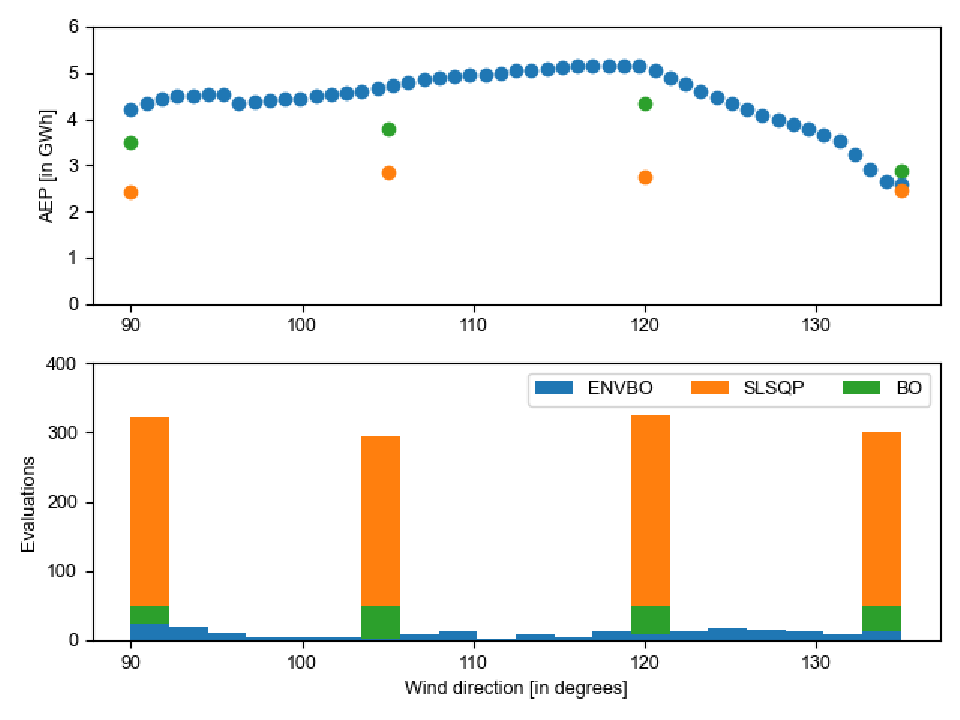}
    \caption{Annual energy production and number of function evaluations conditional on the wind direction. ENVBO (Algorithm~\ref{alg:env-bo}) is benchmarked against SLSQP and BO (Algorithm~\ref{alg:bo})}
    \label{fig:simulator-3}
\end{figure}

While the results in Figure~\ref{fig:simulator-2} show that ENVBO outperforms BO and SLSQP in almost all instances, they do not take into account the different numbers of function evaluations required by each algorithm. Figure~\ref{fig:simulator-3} plots the annual energy production and the number of function evaluations against the wind direction. The lower plot shows that SLSQP uses the most function evaluations by far with $323$, $294$, $324$ and $301$ evaluations for the four wind directions---a total budget of $1,152$ function evaluations---, whereas ENVBO and BO both use $200$ function evaluations in total. BO divides this budget equally over the four wind directions allocating $50$ evaluations per wind direction, while they are distributed via a random walk for ENVBO. Each of ENVBO's bins of the lower plot only contains $2$ to $24$ function evaluations. Compared to SLSQP both Bayesian optimisation algorithms perform much more sample-efficient and use less than $20\%$ of its evaluation budget. ENVBO uses less than half the evaluation for the four wind directions compared to BO but it outperforms BO for three of the four wind directions as shown in the upper plot. This shows the advantage of a global surrogate model that is fit to all controllable and environmental variables. The model uses all available information and is capable of modelling the effect of the environmental conditions.

Although the improved performance of ENVBO compared to BO and SLSQP is already valuable, the main advantage lies in ENVBO's capability to predict a solution for each possible wind direction. This is illustrated by the $51$ solutions given for ENVBO in the upper plot of Figure~\ref{fig:simulator-3}. SLSQP and BO can only give solutions for the specific wind direction they were run for. To achieve similar results, SLSQP and BO need to be run for each wind direction again which would multiply the required function evaluations many times. ENVBO uses the available function evaluation budget much more effectively and is a more sample-efficient and cost-effective approach.

Overall, ENVBO finds solutions over the whole range of the environmental variable, while in most cases performing up to $88\%$ better than algorithms that focus on one fixed environmental variable at a time. This is even true for ranges of the environmental variable that are only explored briefly as the algorithm learns the effect of the environmental variable from adjacent areas. Furthermore, ENVBO uses only a small fraction of the evaluation budget of the two benchmarks making it particularly beneficial for the optimisation of computer simulators and physical experiments that can be very expensive to run in engineering.

\section{Conclusion}\label{sec:conclusion}
When optimising physical experiments, oftentimes not all variables can be fully controlled, or interest lies in finding not one global optimum but one optimum for each value of a certain variable---essentially a function that maps environmental variable values to optimal values for the controllable parameters. In the past, Bayesian optimisation was used predominantly to find global optima. This study extends the Bayesian optimisation algorithm to situations with changing environmental conditions that cannot be controlled. The proposed method fits a global surrogate model over all controllable and environmental variables and uses measurements of the environmental variables to conditionally optimise the acquisition function with regard to the controllable parameters. This conditional optimisation enables finding a model with a posterior predictive mean that provides close to optimal values of the controllable parameters for any values of the uncontrollable variables. With the original Bayesian optimisation algorithm, the uncontrollable variables are assumed fixed or disregarded entirely. Achieving similar results requires repeating the optimisation process for various fixed values for the uncontrollable variables and interpolating between the found optima. Thus, the proposed approach is more sample-efficient as it uses all available information about the objective function in one optimisation run.

This study empirically investigated the properties of the modified Bayesian optimisation strategy from Algorithm~\ref{alg:env-bo} outlined in the previous paragraph on two synthetic test functions---the two-dimensional Levy function and the six-dimensional Hartmann function. The investigation showed that the algorithm manages to solve problems with added noise and uncontrollable variables with high and low variability. When solving problems with more than one environmental variable, it has to be ensured that the final Gaussian process model is not used for extrapolation as this will result in biased solutions. Additionally, it was found that uncontrollable variables with large fluctuations require more function evaluations than uncontrollable variables that fluctuate less. In general, higher fluctuation results in a larger effective parameter domain than lower fluctuation. When modelling, it is intuitive that larger areas require more observations---and thus information---than smaller areas to achieve identical results, assuming that all other properties are comparable.

ENVBO---an implementation of the proposed algorithm within the Python package NUBO \citep{Diessner2023}---was applied to a wind farm simulator with the objective to place four wind turbines within an area with complex underlying terrain to find positions that maximise the annual power generation for different wind directions. The results were compared to two benchmarks---regular Bayesian optimisation via NUBO and the SLSQP algorithm via the SciPy package \citep{SciPy2020}---showing up to $88\%$ better performance in all, but one case across the whole range of possible wind directions while keeping function evaluations and thus costs low. ENVBO presents itself as a sample-efficient and cost-effective approach for the optimisation of expensive experiments and simulators with uncontrollable environmental conditions.

\section*{Acknowledgments}

The work has been supported by the Engineering and Physical Sciences Research Council (EPSRC) under grant number EP/T020946/1 and the EPSRC Center for Doctoral Training in Cloud Computing for Big Data under grant number EP/L015358/1.

\bibliographystyle{unsrt}  
\bibliography{references}

\end{document}